%% file: main.tex
\documentclass[10pt,letterpaper]{article}
\PassOptionsToPackage{table}{xcolor}
\PassOptionsToPackage{no-math}{fontspec}
\usepackage{alibaba-report}
\usepackage{c2ft-paper}
\input{metadata}

\begin{document}

\maketitle

\begingroup
\renewcommand{\thefootnote}{\fnsymbol{footnote}}
\footnotetext{%
  $^{*}$Equal contribution. \quad
  $^{\text{\Letter}}$Corresponding authors: cuibenlei.cbl@alibaba-inc.com, honghaiwen.hhw@alibaba-inc.com.
}
\endgroup

\input{sec/0_abstract}
\input{sec/1_introduction}
\input{sec/2_related_works}
\input{sec/3_method}
\input{sec/4_experiments}
\input{sec/5_conclusion}

\bibliographystyle{plainnat}
{
\small
\bibliography{reference-typeset}
}

\end{document}

%% file: metadata.tex
\title{\method: Enhancing Fine-Grained Perception in\\
MLLMs via Confuse-then-Contrast Fine-Tuning}

\author{%
Shaoxuan He$^{\text{*}}$ \quad
Benlei Cui$^{\text{*} \text{\Letter}}$ \quad
Shikai Qiu \quad
Yuwen Zhai \\
Hui Xue \quad
Longtao Huang \quad
Jingqun Tang \quad
Haiwen Hong$^{\text{\Letter}}$ \\[0.3cm]
{\reportaffiliationfont\mdseries Yuvion Team, Alibaba Group}
}

\newcommand{\shortpapertitle}{C2FT: Enhancing Fine-Grained Perception in MLLMs via Confuse-then-Contrast Fine-Tuning}
\renewcommand{\authorcontributions}{}
\draftfalse
\AtBeginDocument{\hypersetup{%
  pdfauthor={Shaoxuan He, Benlei Cui, Shikai Qiu, Yuwen Zhai, Hui Xue, Longtao Huang, Jingqun Tang, Haiwen Hong}}}

%% file: sec/0_abstract.tex
\begin{abstract}
Multimodal Large Language Models (MLLMs) have demonstrated remarkable capabilities, yet they frequently struggle with fine-grained visual perception, often suffering from hallucinations when faced with visually similar but semantically distinct instances. A major underlying cause is the conventional single-sample Supervised Fine-Tuning (SFT) paradigm, which isolates training instances and limits the model's ability to explicitly learn subtle discriminative boundaries. To address this, we propose \textbf{\method}, a novel \textbf{Confuse-then-Contrast Fine-Tuning} framework that enhances fine-grained perception by shifting from independent supervision to joint supervision.
Specifically, \method assesses the MLLM's internal uncertainty to dynamically mine model-specific semantic confusions, subsequently constructing challenging multi-image groups comprising hard positive and hard negative samples. By interleaving these samples into a unified prompt, our approach forces the MLLM's internal attention mechanisms to cross-reference images and explicitly capture localized visual discrepancies. Extensive experiments on both Fine-Grained Visual Classification (FGVC) and visual question answering (VQA) benchmarks demonstrate the effectiveness of \method. For instance, on the Qwen3-VL-4B model, our approach achieves an average improvements of 3.87\% and 4.35\% points over standard SFT and GRPO baselines, respectively. Project is available at: \url{https://dreamer-hsx.github.io/c2ft-project}.
\end{abstract}

%% file: sec/1_introduction.tex
\section{Introduction}

Multimodal Large Language Models (MLLMs) have achieved remarkable progress across a broad range of vision-language tasks through large-scale pre-training~\cite{comanici2025gemini,hurst2024gpt,deitke2025molmo,bai2025qwen3,guo2025seed1}, demonstrating strong global semantic understanding capabilities. They can recognize common objects, interpret complex scenes, and generate fluent language responses conditioned on visual inputs. Despite these advances, their visual perception remains largely coarse-grained~\cite{fu2024blink,rahmanzadehgervi2024vision,zhang2024visually}. In tasks requiring precise visual discrimination~\cite{zhang2024mme,geigle2024african}, such as spatial reasoning~\cite{yang2025thinking,chen2024spatialvlm,liu2025can,marsili2025visual}, fine-grained instance recognition~\cite{wei2021fine,shu2023learning}, and localized detail attribution~\cite{hirsch2025laquer,wu2023v}, even advanced MLLMs often hallucinate or rely on superficial cues~\cite{li2023evaluating,tong2024eyes,ye2024beaf}. This limitation prevents them from capturing subtle structural, textural, or spatial differences, which are essential for distinguishing visually similar instances and resolving fine-grained visual ambiguities.

\begin{figure}
    \centering
    \includegraphics[width=\linewidth]{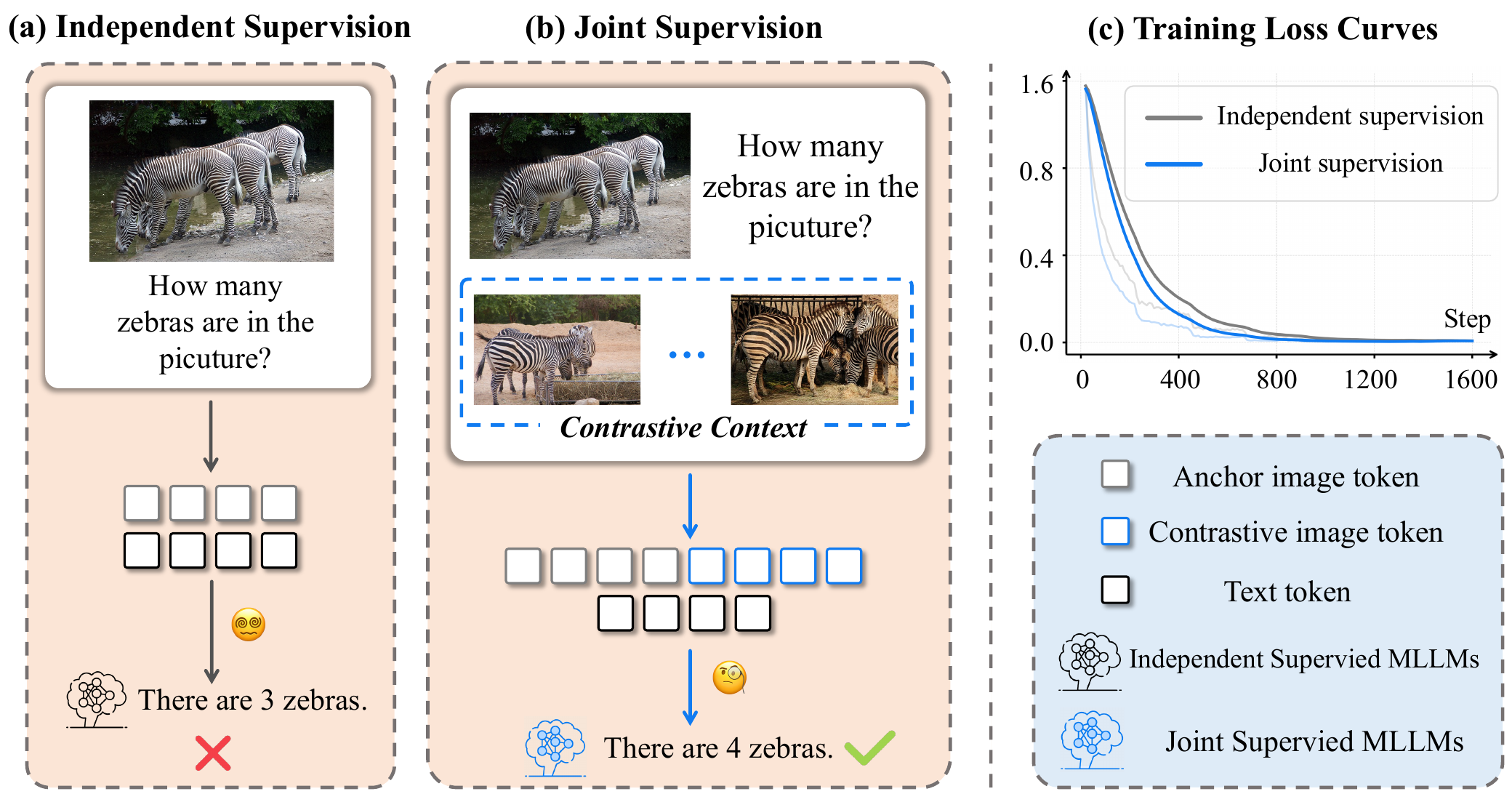}
    \caption{\textbf{Comparison between Independent and Joint Supervision paradigms for MLLMs.} (a) Under standard independent supervision, MLLMs often fail on complex queries due to a lack of comparative context. (b) By incorporating auxiliary images as a contrastive context pool, our joint supervision explicitly encourages the model to differentiate fine-grained features. (c) Joint supervision achieves faster convergence than the independent baseline.}
    \label{fig:teaser}
\end{figure}

Recent studies have primarily sought to mitigate this issue through specialized supervised fine-tuning~\cite{li2025vocot,muennighoff2025s1,zelikman2022star} or reinforcement learning methods~\cite{ouyang2022training,rafailov2023direct} such as DPO. Although effective to some extent, these approaches often require expensive fine-grained annotations~\cite{zhang2024visually,he2025analyzing,shi2025enhancing} or manually crafted Chain-of-Thought (CoT) rationales~\cite{luo2025unlocking,xu2025llava,thawakar2025llamav}. More importantly, most existing approaches follow an independent supervision paradigm, in which each training sample is optimized in isolation on the input side. This isolated training scheme provides limited contrastive supervision and thus makes it difficult for the model to identify subtle instance-level differences. As shown in \cref{fig:teaser}(a), models tend to form coarse semantic associations rather than sharp discriminative boundaries, leaving them vulnerable to confusing fine-grained details.

In this paper, we propose \textbf{\method}, a \textbf{Confuse-then-Contrast Fine-Tuning} framework to enhance fine-grained perception in MLLMs. Inspired by the success of contrastive approaches~\cite{deng2019arcface,schroff2015facenet} in traditional vision tasks, we integrate a comparative mechanism directly into the MLLM with input side joint supervision.
Specifically, \method first identifies samples that are likely to confuse the model based on output likelihoods and visual feature similarities. It then constructs hard negative pairs with similar visual appearances but different answers, as well as hard positive pairs with diverse appearances but similar answers. These samples are combined into unified multi-image instructions, enabling cross-image attention and encouraging the model to explicitly compare instances. Moreover, to make multi-image training more robust, we randomly shuffle image orders to reduce positional bias and mix the constructed multi-image data with single-image samples to mitigate catastrophic forgetting~\cite{yu2024rlhf,li2025think}. As shown in \cref{fig:teaser}(b), this process encourages MLLMs to learn fine-grained details beyond standard independent supervision.

To validate the effectiveness of \method, we conduct extensive experiments across diverse fine-grained multimodal benchmarks, encompassing both FGVC and VQA tasks. Empirical results demonstrate that our approach outperforms standard SFT and GRPO baselines. Our method achieves an overall average score of 82.95\% on Qwen3-VL-4B, exceeding the SFT and GRPO baselines by 3.87 and 4.35 points, respectively; on Gemma-4-E2B, the corresponding gains are 2.72 and 3.56 points. Furthermore, we observe a promising scaling trend: \textit{expanding the pool of interleaved contrastive samples consistently enhances fine-grained perceptual performance}. These findings further support our core motivation that enabling MLLMs to explicitly learn from model-specific confusions and cross-image comparisons is crucial for resolving subtle visual ambiguities. Our contributions are threefold:

\begin{itemize}[leftmargin=*]
    \item We identify the limitations of isolated single supervised training for fine-grained multimodal perception, highlighting its lack of contrastive supervision for distinguishing visually similar but semantically different instances.
    \item We propose \method, a Confuse-then-Contrast Fine-Tuning framework. This approach dynamically mines model-specific hard positives and hard negatives and organizes them into multi-image instructions to enforce explicit cross-image attention and discriminative learning.
    \item We conduct extensive experiments on FGVC and VQA benchmarks, showing that \method outperforms standard SFT and GRPO baselines across different model scales, while also revealing a clear scaling trend as the number of interleaved contrastive samples increases.
\end{itemize}

%% file: sec/2_related_works.tex
\section{Related Work}

\subsection{Fine-grained Perception of MLLMs}
MLLMs have been widely deployed across a broad spectrum of real-world applications~\cite{cui2026metavideoagent,he2025neighboring,he2026dynamics,he2026tc,qiu2026yuvion}, motivating increasing attention to their fine-grained visual perception capabilities. One line of work adopts training-free strategies~\cite{peng2025survey}, such as prompt engineering~\cite{mohanty2025future}, or attention-based feature extraction~\cite{huang2022cross,chen2024fet}, to better elicit fine-grained visual cues without additional model updates. While computationally efficient, these methods often lack sufficient discriminative supervision, making it difficult to reliably separate visually similar categories. Another line of work employs knowledge distillation (KD)~\cite{he2025analyzing,hu2024visual,sarch2024vlm} to transfer task-specific fine-grained skills from powerful teacher models to more efficient student networks. Although KD significantly reduces computational overhead~\cite{shao2023data,jang2025vl2lite}, it primarily aligns intermediate features or output logits, often missing the critical fine-grained alignment between localized visual objects and specific categories~\cite{he2025analyzing,lin2025perspective}.

Beyond direct perception enhancement, chain-of-thought (CoT) reasoning has been introduced to encourage MLLMs to decompose fine-grained recognition into intermediate visual evidence before predicting the final category~\cite{zhu2026can,xu2025llava}. By providing explicit reasoning traces, CoT can improve interpretability and visual grounding. While explicit reasoning traces can improve interpretability and visual grounding, their effectiveness depends heavily on the quality of the reasoning data, which is difficult to obtain at scale for fine-grained tasks~\cite{yang2025new,xu2024fineparser}. Reinforcement learning~\cite{he2026fine,ding2026tikart,liu2025visual} offers another way to improve visual alignment by rewarding image-grounded responses. However, their application to fine-grained visual perception usually requires specially designed reward signals and suffers from reward hacking~\cite{skalse2022defining,fu2025reward}.

\subsection{Visual Contrastive Training}
The concept of visual contrastive training refers to learning visual representations by encouraging semantically related inputs to be closer in the feature space while pushing unrelated ones farther apart~\cite{chen2020simple}. This paradigm has shown strong effectiveness in self-supervised learning~\cite{he2020momentum,radford2021learning,liang2022mind} where different augmented views of the same image are treated as positive pairs and other samples serve as negatives. Beyond such view-level objectives~\cite{tian2020contrastive}, prior work has also investigated explicit cross-image contrast~\cite{ding2024joint,yao2022image,dunlap2024describing}, which enables models to capture finer-grained distinctions and relational cues across instances. In multimodal settings, contrastive vision-language alignment~\cite{yang2022vision,alayrac2022flamingo,lin2025comparison} has significantly advanced MLLMs, enhancing visual discrimination and relational reasoning. Recently, contrastive principles have emerged as data construction strategies for instruction tuning~\cite{pan2026through,ma2024c3l,jiao2025img}, leveraging hard negatives to refine MLLM comparative perception and fine-grained judgment. However, how to effectively perform visual contrasting for MLLMs without external knowledge remains underexplored. To address this, \method introduces a confuse-then-contrast paradigm that dynamically mines hard contrastive samples, providing explicit joint supervision for fine-grained perception.

%% file: sec/3_method.tex
\section{Method}
\begin{figure}
    \centering
    \includegraphics[width=\linewidth]{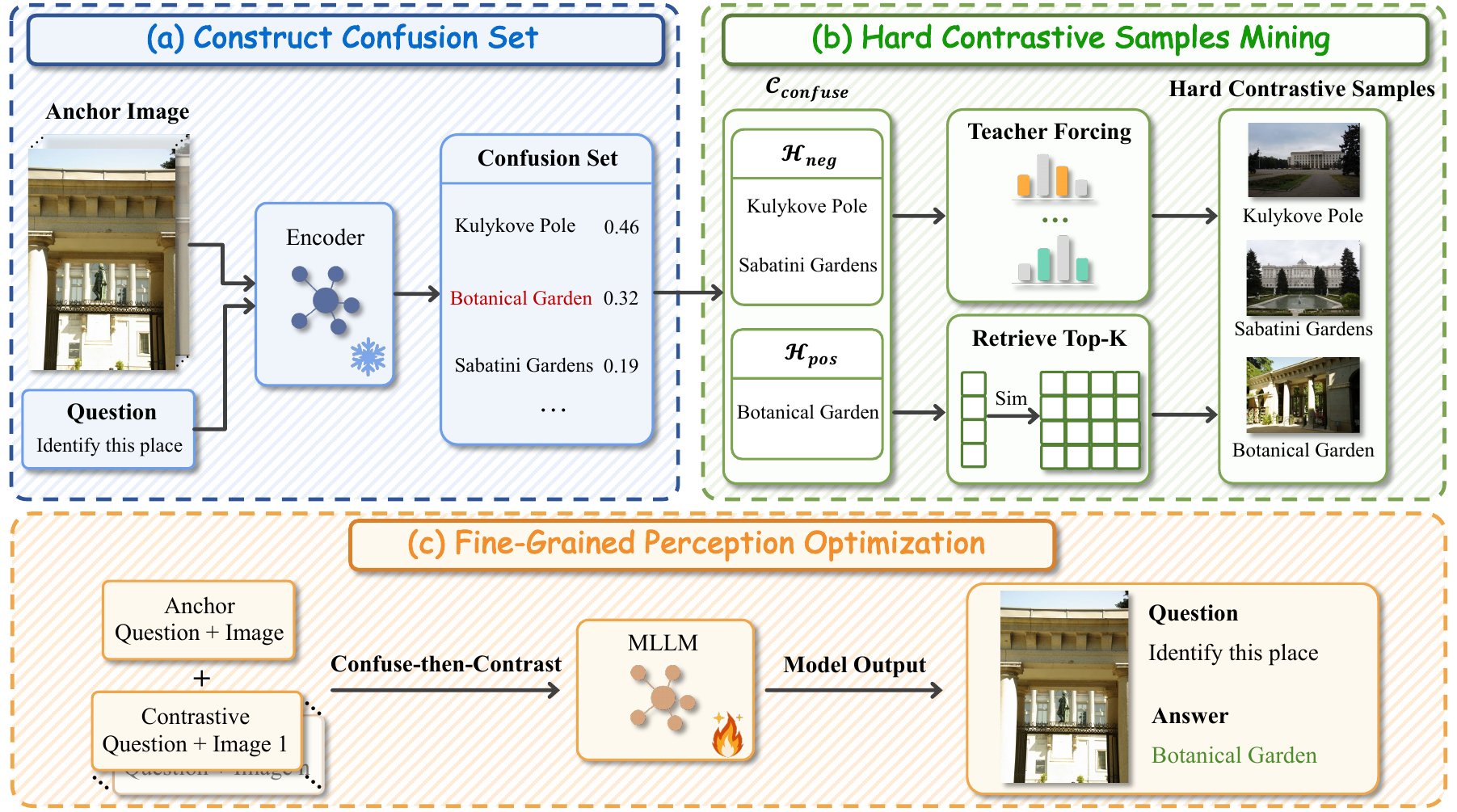}
    \caption{\textbf{Overview of \method.} (a) \method first constructs a dynamic semantic confusion set for the given anchor samples. (b) Then hard contrastive samples are mined: positives are retrieved directly through feature similarities, whereas negatives are verified via Teacher Forcing generation probabilities. (c) The assembled multi-image context is joint optimized fine-grained visual perception.}
    \label{fig:framework}
\end{figure}

Our method, \method, enhances fine-grained perception in MLLMs through a confuse-then-contrast training paradigm. We first introduce a mining module that leverages the model's own uncertainty to identify visually or semantically confusable categories, and uses a strong encoder~\cite{zhang2025qwen3,simeoni2025cijo} to retrieve hard positive and negative samples.
We then adopt a progressive anti-shortcut training strategy that combines shuffled multi-image contrastive training with mixed-format supervision. This progressive design encourages the model to capture discriminative fine-grained visual cues, while mitigating positional shortcuts and preserving its basic recognition capability. \cref{fig:framework} provides an overview of the proposed \method framework.

\subsection{Preliminary}
\noindent\textbf{Joint Attention over Multi-Image.} 
Modern MLLMs typically encode each input image into a sequence of visual tokens and concatenate them with textual tokens before feeding them into a Transformer-based language model. Given multiple images, their visual tokens are placed in a shared token sequence, allowing the self-attention layers to compute dependencies not only between text and image tokens, but also across visual tokens from different images. We refer to this mechanism as joint attention over multiple images.

Formally, let $\mathbf{V}_i = \{v_i^1, \ldots, v_i^{n_i}\}$ denote the visual tokens of the $i$-th image, and let $\mathbf{T}$ denote the textual tokens. The input sequence can be written as
\[
\mathbf{X} = [\mathbf{V}_1; \mathbf{V}_2; \ldots; \mathbf{V}_K; \mathbf{T}],
\]
where $K$ is the number of images. Standard self-attention is then computed over the entire sequence:
\[
\mathrm{Attn}(\mathbf{Q}, \mathbf{K}, \mathbf{V}) =
\mathrm{softmax}\left(\frac{\mathbf{Q}\mathbf{K}^{\top}}{\sqrt{d}}\right)\mathbf{V}.
\]
Since all image tokens participate in the same attention operation, tokens from one image can directly attend to tokens from another image. This provides a structural basis for cross-image comparison and enables the model to relate visually similar regions, objects, or attributes across different images.

\noindent\textbf{Comparative Supervision for Discrimination.}
Fine-grained perception is inherently comparative, as visually similar samples often differ only in subtle attributes~\cite{deng2019arcface}. Standard single-sample SFT maximizes $p_{\theta}(y \mid I, q)$ for each image-text pair independently, which provides limited discriminative signal. Comparative supervision addresses this limitation by introducing a positive image $I^{+}$ and a set of hard negatives $\mathcal{I}^{-}=\{I_1^{-}, \ldots, I_m^{-}\}$. The model is encouraged to rank the matched pair above mismatched ones:
\[
s_{\theta}(I^{+}, q, y) > s_{\theta}(I_j^{-}, q, y), \quad \forall I_j^{-}\in \mathcal{I}^{-},
\]
where $s_{\theta}$ denotes the model's compatibility score. By contrasting visually similar candidates, this supervision makes discriminative cues more salient and encourages the model to exploit multi-image joint attention for fine-grained reasoning.

\subsection{Confuse-then-Contrast Mining}
The Confuse-then-Contrast Mining (C2M) module dynamically constructs challenging multi-image exemplars by coupling the MLLM's intrinsic semantic uncertainty with fine-grained visual retrieval. Operating on-the-fly at the end of each training epoch, C2M proceeds in two sequential steps: model-centric confusion estimation and fine-grained visual contrastive retrieval.

\noindent\textbf{Model-Centric Confusion Estimation.}
Traditional hard negative mining typically relies on static visual similarities, which often fails to capture the actual semantic weaknesses of MLLMs dynamically. To address this, we propose a model-centric confusion estimation strategy based on feature-level semantic retrieval. 

Prior to training, we pre-compute and construct an offline feature bank. This includes extracting the text embeddings for all ground-truth (GT) answers and query texts, as well as the visual embeddings for all images using a frozen pre-trained encoder. At the end of each epoch, we then activate the dynamic estimation mechanism.

Given an anchor sample $x_a$ with its ground-truth $y_a$, if the model yields a wrong answer, we extract the text embedding of this generated incorrect prediction, denoted as $e_{pred}$. To quantify the model's current semantic confusion, we calculate the similarity (e.g., cosine similarity) between $e_{pred}$ and the pre-computed GT text embeddings $e_{gt}^c$ of all other candidate categories $c \in \mathcal{C}$:
\begin{equation}
    S_{confuse}(c \mid x_a) = \text{Sim}(e_{pred}, e_{gt}^c).
\end{equation}
By sorting these candidate categories in descending order based on their similarity scores $S_{confuse}$, we obtain a ranked confusion set $\mathcal{C}_{confuse}$. The categories ranked at the top of $\mathcal{C}_{confuse}$ represent the concepts that the current model is most heavily confusing with the anchor, thereby serving as highly targeted semantic anchors for the subsequent fine-grained visual retrieval.

\noindent\textbf{Fine-Grained Visual Contrastive Retrieval.}
Having identified the dynamic semantic confusion set $\mathcal{C}_{confuse}$, we proceed to construct a highly challenging multi-image context. To ensure the contextual instances provide optimal contrastive difficulty, we apply distinct retrieval and verification strategies for hard negative and hard positive samples.

For \textbf{hard negative samples}, we draw candidate images from the classes in $\mathcal{C}_{confuse}$. To verify whether a specific candidate sample genuinely induces cognitive confusion for the MLLM, we employ a Teacher Forcing verification step. For a candidate negative sample $x_{neg}$ belonging to a confused class $c \in \mathcal{C}_{confuse}$, we compute the model's generation probability of the anchor's ground-truth label $y_a$, denoted as $P_{TF}(y_a \mid x_{neg})$. We evaluate the margin between this confusion probability and the probability of predicting its true label $P_{TF}(c \mid x_{neg})$. If the confusion probability is critically high, which indicates a small margin with the ground truth, then $x_{neg}$ is explicitly verified and collected into the hard negative set $\mathcal{H}_{neg}$:
\begin{equation}
    \mathcal{H}_{neg} = \left\{ x_{neg} \in \mathcal{D}_{c} \mid c \in \mathcal{C}_{confuse}, \; \lvert P_{TF}(y_a \mid x_{neg}) -P_{TF}(c \mid x_{neg}) \rvert < \tau \right\},
    \label{eq:similarity_threshold}
\end{equation}
where $\tau$ is a predefined threshold (see ablation study in \cref{sec:threshold}).

For \textbf{hard positive samples}, which are distinct instances belonging to the anchor's ground-truth class $y_a$, we bypass the computationally expensive Teacher Forcing step. Instead, we perform an efficient multi-modal feature retrieval. We retrieve candidate sample $x_{pos}$ using the anchor's pre-computed image and text representation, denoted as $e_{x_a}^i$ and $e_{x_a}^t$ respectively. To ensure the selected positives are visually and semantically relevant, we enforce a constraint that the retrieved samples $x_{pos}$ must be highly aligned with the anchor's image and text embedding. The hard positive set $\mathcal{H}_{pos}$ is thus constructed by selecting the Top-$K$ samples that maximize this joint similarity:
\begin{equation}
    \mathcal{H}_{pos} = \mathop{\text{Top-}K}_{x_{pos} \in \mathcal{D}_{y_a} \setminus \{x_a\}} \Big( \frac{\text{Sim}(e_{x_{pos}}^i, e_{x_a}^i) + \text{Sim}(e_{x_{pos}}^t, e_{x_a}^t)}{2} \Big)
\end{equation}

Finally, the anchor $x_a$, the verified hard positives $\mathcal{H}_{pos}$, and the hard negatives $\mathcal{H}_{neg}$ are concatenated into a unified contrastive group. By feeding this assembled context into the MLLM for joint supervised training, we explicitly implement the cross-image attention mechanisms. This paradigm forces the model to perform fine-grained visual comparisons within the context, comprehensively bridging the attention across instances to overcome its previous semantic bottlenecks.

\subsection{Progressive Anti-Shortcut Training}
In practice, we found that training the MLLM only with a fixed multi-image format can introduce undesirable shortcuts (See \cref{sec:ablation_training}). Specifically, the model tends to generate multiple semantically similar answers in a single response, thereby increasing the chance that the correct answer is included without truly resolving the fine-grained distinction. Moreover, excessive exposure to multi-image instructions weakens the model's instruction-following ability. To address these issues, we propose a progressive anti-shortcut training strategy that combines order-agnostic multi-image contrastive learning with mixed-format supervision.

Given an anchor image $x_a$, we first collect the hard positives and hard negatives mined by the C2M module. To prevent the model from relying on fixed positions or formatting patterns, we merge these samples into a unified context pool and randomly shuffle their order:
\begin{equation}
X_{\mathrm{context}} = \mathrm{Shuffle}(\mathcal{H}_{\mathrm{pos}} \cup \mathcal{H}_{\mathrm{neg}}).
\end{equation}
The shuffled context images are then concatenated with the anchor image in the multimodal prompt. The model is optimized with the cross-entropy loss:
\begin{equation}
\mathcal{L}_{\mathrm{contrast}}
=
\mathbb{E}
\left[
\mathcal{L}_{\mathrm{CE}}
\left(
\theta;
[x_a, X_{\mathrm{context}}],
[y_a, Y_{\mathrm{context}}]
\right)
\right],
\end{equation}
where $Y_{\mathrm{context}}$ denotes the ground-truth labels corresponding to the contextual images. Since the positions of hard positives and negatives are randomized, the model cannot infer the answer from image order or template structure, and is instead encouraged to compare fine-grained visual evidence.

To further preserve the model's single-image recognition and instruction-following capabilities, we adopt a mixed-format training objective. Each training batch contains a proportion $\rho$ of standard single-image samples and $1-\rho$ of mined multi-image contrastive samples. The overall objective is:
\begin{equation}
\mathcal{L}_{\mathrm{total}}
=
\rho \,
\mathbb{E}_{(x,y)\sim \mathcal{B}_{\mathrm{single}}}
\left[
\mathcal{L}_{\mathrm{CE}}(\theta; x, y)
\right]
+
(1-\rho) \,
\mathbb{E}_{(X,Y)\sim \mathcal{B}_{\mathrm{multi}}}
\left[
\mathcal{L}_{\mathrm{contrast}}
\right].
\end{equation}
In practice, we set $\rho=0.3$ to balance the instruction following ability and the multi-image contrastive objective.
By combining shuffled hard exemplars with mixed-format supervision, the proposed training strategy encourages discriminative fine-grained reasoning while maintaining robust general instruction-following behavior.

%% file: sec/4_experiments.tex
\section{Experiments}

\subsection{Experimental Settings}
\noindent\textbf{Datasets.}
We evaluate our proposed \method across two primary categories requiring fine-grained perception: Fine-Grained Visual Classification (FGVC) and Visual Question Answering (VQA). For FGVC, we utilize FGVC-Aircraft~\cite{maji2013fine}, Oxford-IIIT Pet (Pet-37)~\cite{parkhi2012cats}, Stanford Cars (Car-196)~\cite{krause20133d}, and the Google Landmarks Dataset~\cite{weyand2020google}. For VQA tasks, which contain a more diverse query and answer format, we employ TallyQA~\cite{acharya2019tallyqa} and GQA~\cite{hudson2019gqa}. TallyQA comprises 287K complex counting questions across 165K images with numerical answers up to 15, and GQA is a dataset consisting of 22M challenging compositional questions over real-world images that evaluates spatial understanding, inherently demanding strong fine-grained visual perception capabilities from the models.

\noindent\textbf{Baselines and metrics.}
We evaluate our method against two groups of baselines. First, we compare with leading open-source MLLMs of varying scales~\cite{li2024llava, laurenccon2023obelics}, including the InternVL~\cite{chen2024internvl} series and Qwen2.5-VL~\cite{bai2025qwen3} series. Second, to perform a rigorous internal ablation, we apply \method to foundational models (Qwen3-VL-4B and Gemma-4-E2B~\cite{team2024gemma}) and compare the results against their Base, SFT, and GRPO counterparts. Following standard evaluation protocols, we report the Top-1 Accuracy for all individual datasets and calculate the overall average score.

\noindent\textbf{Implementation details.}
We utilize Qwen3-VL-4B and Gemma-4-E2B as our base models. For all of our experiments, we adopted the same learning rate as the baseline, utilized
the AdamW optimizer, and trained the models until convergence. All models are trained using full-parameter fine-tuning, and the maximum sequence length is capped at 16384 tokens.

\begin{table*}[t]
\centering
\caption{\textbf{Quantitative comparisons on FGVC and VQA benchmarks.} We report the quantitative results on both FGVC and VQA datasets. The best results are \textbf{bolded} and the second best are \underline{underlined}. Aircraft: FGVC-Aircraft dataset. GLD: Google Landmarks Dataset.}
\setlength{\tabcolsep}{8pt}
\renewcommand{\arraystretch}{1.15}

\begin{adjustbox}{max width=\textwidth}
\begin{tabular}{lccccccc}
\toprule
\multirow{2}{*}{Models} 
& \multicolumn{4}{c}{FGVC} 
& \multicolumn{2}{c}{VQA} 
& \multirow{2}{*}{AVG} \\
\cmidrule(lr){2-5}
\cmidrule(lr){6-7}
& Aircraft & Pet-37 & Car-196 & GLD & TallyQA & GQA & \\
\midrule

\rowcolor{gray!20}
\multicolumn{8}{c}{\textit{Comparison with other MLLMs}} \\
\midrule

LLaVA-Onevision-7B             & 31.51 & 51.23 & 46.77 & 34.84 & 42.76 & 48.95 & 42.68 \\
Idefics3-LLAMA3-8B   & 32.76 & 55.82 & 50.49 & 37.81 & 45.63 & 51.74 & 45.71 \\
InternVL2.5-4B         & 29.45 & 43.16 & 51.04 & 33.42 & 39.70 & 46.93 & 40.62 \\
InternVL2.5-8B    & 34.96 & 56.32 & 53.17 & 38.90 & 43.72 & 51.58 & 46.44 \\
Qwen2.5-VL-3B       & 41.48 & 56.92 & 57.69 & 42.83 & 49.62 & 54.87 & 50.57 \\
Qwen2.5-VL-7B       & 44.94 & 59.85 & 60.77 & 48.53 & 54.50 & 58.31 & 54.48 \\

\midrule
\rowcolor{gray!20}
\multicolumn{8}{c}{\textit{Comparison among Our Models}} \\
\midrule

DeepPerception-7B                    & 44.24 & 40.90 & 54.14 & 49.35 & - & - & 47.15 \\
No-Thinking-RFT                    & 74.41 & 86.13 & \underline{92.5} & 64.87 & - & - & \underline{79.48} \\

\midrule

Qwen3-VL-4B                    & 47.55 & 59.70 & 63.92 & 53.54 & 59.52 & 61.64 & 57.65 \\
\quad + SFT                 & 84.52 & \textbf{92.23} & 85.72 & \underline{82.31} & \underline{63.96} & \underline{65.74} & 79.08 \\
\quad + GRPO                 & \underline{85.13} & 89.64 & 86.96 & 81.60 & 63.42 & 64.83 & 78.60 \\
\quad + \method                 & \textbf{89.43} & \underline{91.67} & \textbf{93.27} & \textbf{84.72} & \textbf{68.45} & \textbf{70.16} & \textbf{82.95} \\

\midrule

Gemma-4-E2B       & 21.18 & 33.92 & 32.75 & 28.64 & 36.81 & 38.73 & 32.01 \\
\quad + SFT           & 37.39 & \underline{62.68} & \underline{59.86} & \underline{66.72} & 45.25 & \underline{44.81} & \underline{52.79} \\
\quad + GRPO                 & \underline{38.41} & 60.10 & 59.72 & 64.31 & \underline{46.92} & 42.27 & 51.95 \\
\quad + \method                 & \textbf{40.68} & \textbf{65.72} & \textbf{61.59} & \textbf{68.73} & \textbf{48.43} & \textbf{47.91} & \textbf{55.51} \\

\bottomrule
\end{tabular}
\end{adjustbox}

\label{tab:main-results}
\end{table*}

\begin{figure}
    \centering
    \includegraphics[width=\linewidth]{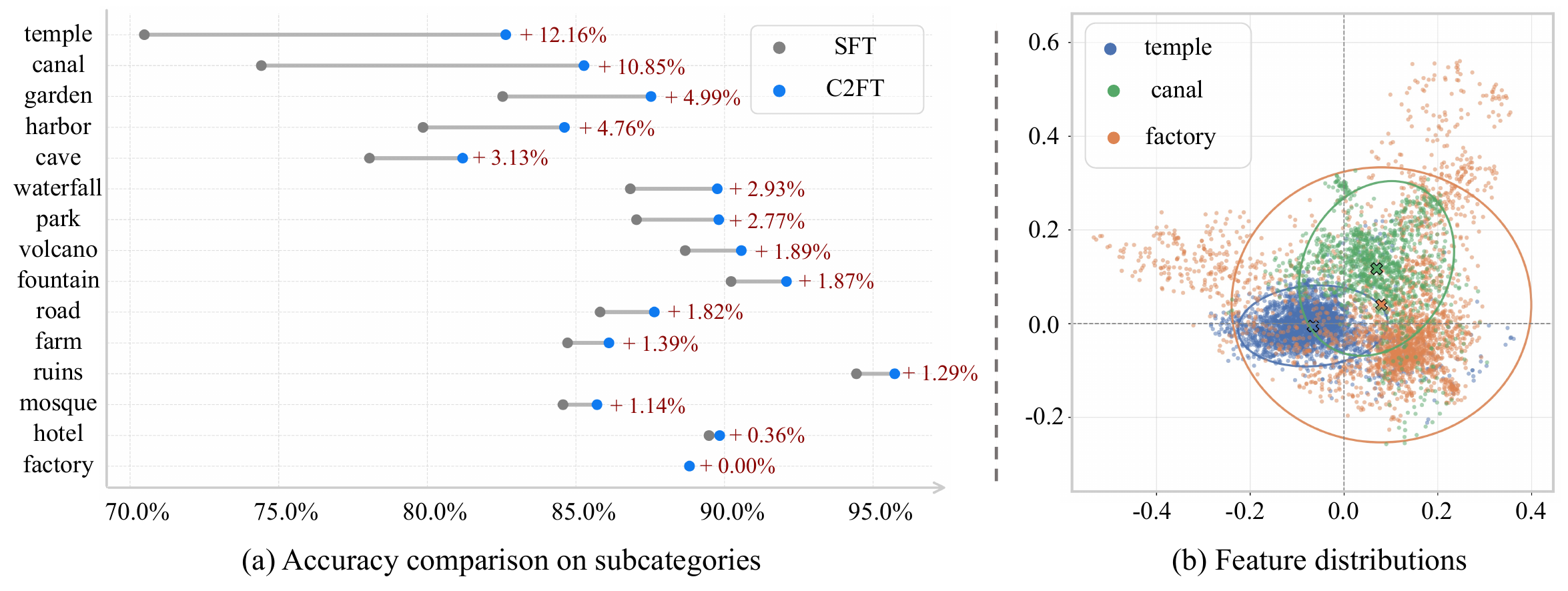}
    \caption{\textbf{Per category analysis on the Google Landmarks dataset.} (a) Accuracy gains of \method over the SFT baseline across categories. (b) PCA visualization of image embeddings. The results demonstrate that classes with more densely clustered and visually entangled feature spaces can yield larger performance improvements.}
    \label{fig:gld_analysis}
\end{figure}

\subsection{Main Results}

\noindent\textbf{Quantitative analysis.} In this section, we train and evaluate our method across FGVC and VQA datasets, with results summarized in \cref{tab:main-results}. \method achieves the best overall average score on both Qwen3-VL-4B and Gemma-4-E2B, raising the score from 57.65 to 82.95 and 32.01 to 55.51, respectively. It also surpasses recent fine-grained perception methods such as DeepPerception-7B~\cite{ma2025deepperception} and No-Thinking-RFT~\cite{li2025think}. Although No-Thinking-RFT shows competitive performance on several FGVC datasets, \method further improves the results on Aircraft, Car-196, and GLD, demonstrating its effectiveness in both fine-grained visual categorization and visual question answering.

Compared with different fine-tuning strategies, \method also shows clear advantages. On Qwen3-VL-4B, it surpasses standard SFT and GRPO by 3.87 and 4.35 average points, respectively. Similar trends are observed on Gemma-4-E2B, where \method outperforms both SFT and GRPO across all benchmarks. These consistent gains across different backbones suggest that introducing input-side comparison among confusing samples helps MLLMs learn more discriminative fine-grained visual cues than isolated single-sample optimization.

\begin{figure}[t]
    \centering
    \includegraphics[width=\linewidth]{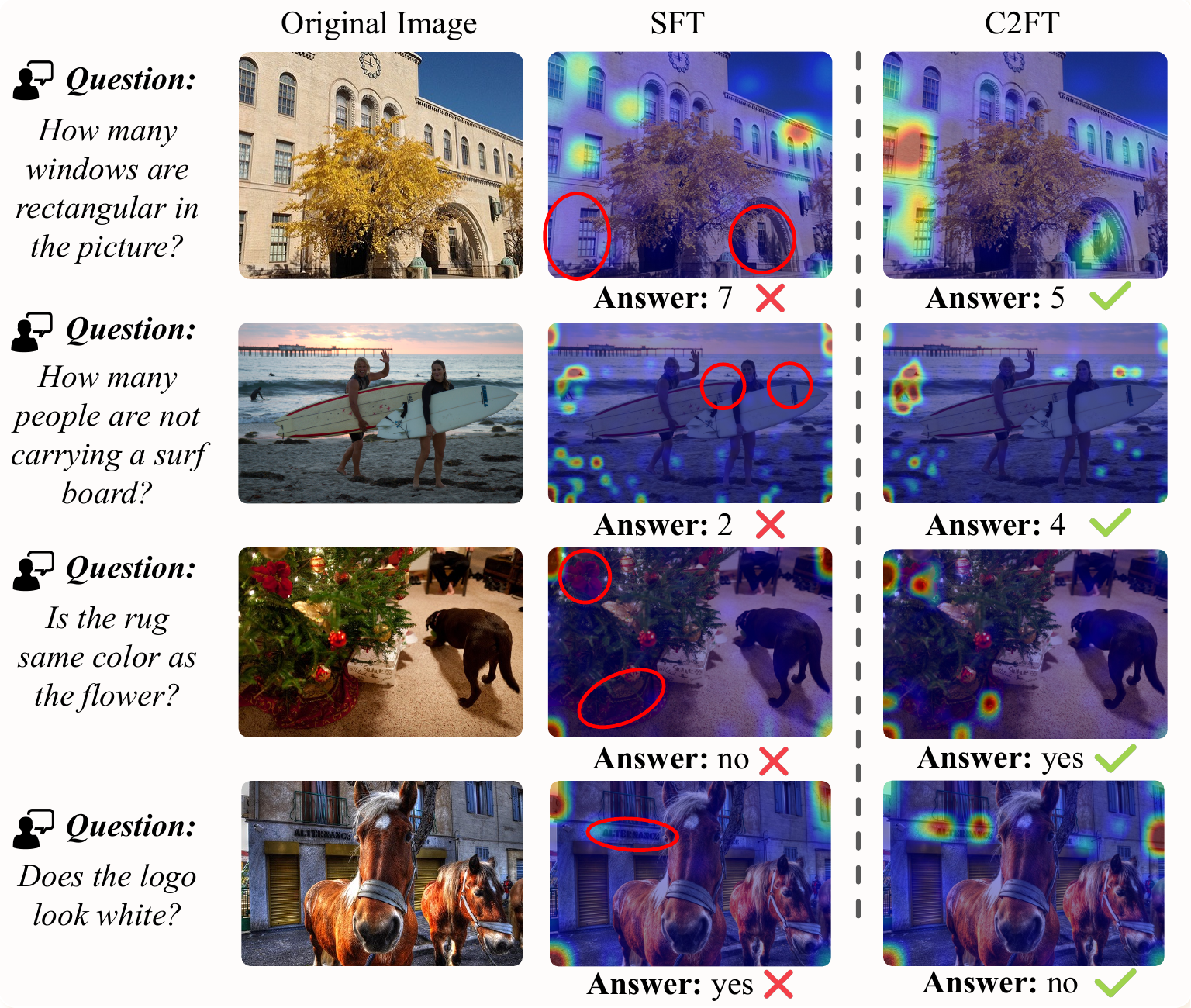}
    \caption{\textbf{Qualitative results on TallyQA and GQA.} Attention maps for the generated answer tokens are visualized, with \textcolor{red}{red circles} highlighting the critical fine-grained features. \method accurately localizes subtle discriminative regions, while the SFT baseline struggles with background distractions, resulting in incorrect predictions.}
    \label{fig:attention_map}
\end{figure}

\noindent\textbf{Qualitative analysis.} \cref{fig:attention_map} compares the attention maps of the standard SFT baseline and \method. The SFT model often attends to irrelevant regions or distractors, leading to incorrect answers in fine-grained counting and attribute comparison tasks. For example, it focuses on invalid windows, people carrying surfboards, or background areas unrelated to the queried objects. In contrast, \method concentrates on the discriminative regions required by the question and produces correct predictions. These results indicate that training with hard contrastive examples helps the model suppress spurious cues and improve fine-grained visual perception.

\noindent\textbf{In-depth analysis on GLD.}
To further investigate how \method enhances fine-grained perception, we conduct a class-level analysis on selected subcategories from the GLD in \cref{fig:gld_analysis}. By correlating performance metrics with feature distributions, we observe a distinct pattern: \method yields the most substantial improvements on categories with densely clustered feature spaces. As shown in \cref{fig:gld_analysis}(b), the embeddings for \textit{temple} and \textit{canal} are heavily entangled, indicating high visual similarity that typically confuses standard MLLMs. Consequently, \method achieves its highest gains on these categories (+12.16\% and +10.85\%, respectively). Conversely, the \textit{factory} category exhibits a widely dispersed feature distribution with less intra-class consistency, resulting in negligible improvement (+0.00\%). This phenomenon validates our core motivation. The confuse-then-contrast mining strategy actively isolates visually ambiguous hard negatives within dense clusters, forcing the model to learn crucial discriminative details exactly where foundational models struggle the most.

\subsection{Ablation Study}

\begin{figure*}[t]
\centering

\begin{minipage}[t]{0.3\linewidth}
\vspace{0pt}
\centering
\small
{\raggedright
\textbf{(a) Ablation on core training strategies.} ``w/o C2M'' denotes replacing C2M with randomized contrastive training; ``w/o PAT'' removes the PAT strategy directly.\par}

\vspace{0.5em}
\resizebox{\linewidth}{!}{
\begin{tabular}{c|cc}
\toprule
Method  & GLD & GQA \\
\midrule
w/o C2M & 81.26 & 65.07 \\
w/o PAT & 72.64 & 51.42 \\
\cellcolor{gray!20} Ours & \cellcolor{gray!20} 84.72 & \cellcolor{gray!20} 70.16 \\
\bottomrule
\end{tabular}
}
\end{minipage}
\hfill
\begin{minipage}[t]{0.38\linewidth}
\vspace{0pt}
\centering
\includegraphics[width=\linewidth]{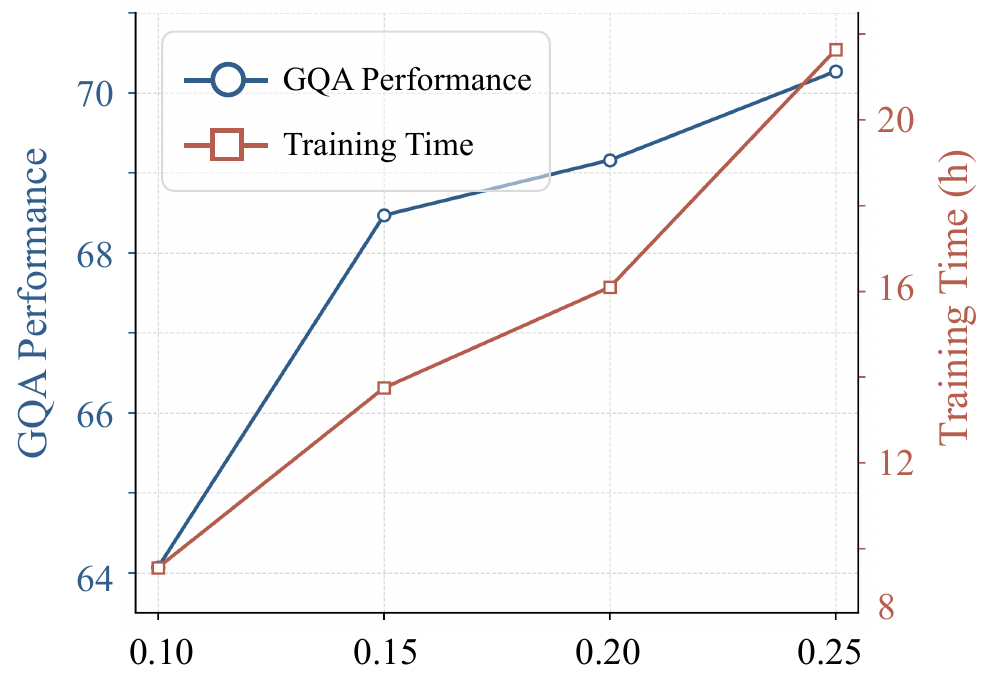}

\small \textbf{(b) Time and effect trade-off across $\tau$.}
\end{minipage}
\hfill
\begin{minipage}[t]{0.3\linewidth}
\vspace{0pt}
\centering
\small
{\raggedright
\textbf{(c) Ablation on threshold $\tau$.} $\tau=0.20$ achieves the best trade-off.\par}

\vspace{0.5em}
\resizebox{\linewidth}{!}{
\begin{tabular}{c|cc}
\toprule
$\tau$ & GLD & GQA \\
\midrule
0.10 & 82.73 & 66.92 \\
0.15 & 83.82 & 69.47 \\
\cellcolor{gray!20} 0.20 & \cellcolor{gray!20} 84.72 & \cellcolor{gray!20} 70.16 \\
0.25 & 83.92 & 71.27 \\
\bottomrule
\end{tabular}
}
\end{minipage}

\caption{\textbf{Ablation results.} (a) Ablation on core training strategies demonstrates the necessity of the C2M and PAT modules. (b) The trade-off between performance and training time across different $\tau$ values. (c) Sensitivity analysis of the threshold $\tau$.}
\label{fig:ablation_all}
\end{figure*}

\noindent\textbf{Effect of core training strategies.}
\label{sec:ablation_training}
To validate the effectiveness of our proposed components, we ablate the C2M and PAT modules, as shown in \cref{fig:ablation_all}(a). Replacing C2M with randomized contrastive training causes a notable performance drop of over 3 and 5 points on GLD and GQA, respectively. Meanwhile, omitting the PAT strategy leads to even more severe degradation across both metrics, underscoring its crucial role in maintaining the model's instruction-following behavior. These results confirm that both training strategies are essential and complementary for maximizing the model's overall performance.

\noindent\textbf{Effect of threshold $\tau$.}
\label{sec:threshold}
We investigate the impact of the threshold $\tau$, which controls the strictness of our hard negative mining. According to \cref{eq:similarity_threshold}, increasing $\tau$ relaxes the margin constraint, thereby allowing a larger quantity of candidate samples to be retrieved into the hard negative set $\mathcal{H}_{neg}$. 

As shown in \cref{fig:ablation_all}(c), performance generally improves as $\tau$ increases, suggesting that a larger pool of hard negatives is beneficial. For example, GQA accuracy steadily increases from 66.92 to 71.27 as $\tau$ grows from 0.1 to 0.25. GLD shows a similar upward trend up to $\tau=0.2$, improving from 82.73 to 84.72, although it slightly drops at $\tau=0.25$; we discuss this anomaly in \cref{sec:limitation}. However, \cref{fig:ablation_all}(b) reveals a critical trade-off between performance and computational cost. Expanding the negative set drastically increases training time, with a particularly steep surge when moving from $\tau$=0.20 to 0.25. Consequently, we adopt $\tau=0.20$ as it achieves the optimal balance between effectiveness and efficiency. This scaling trend validates the effectiveness of \method, demonstrating that exposing the model to more visually ambiguous samples strengthens fine-grained perception.

%% file: sec/5_conclusion.tex
\section{Conclusion}
In this paper, we introduce \method, a Confuse-then-Contrast Fine-Tuning framework designed to enhance the fine-grained visual perception of MLLMs. By dynamically assembling hard positive and negative samples into unified multi-image instructions, \method enforces explicit cross-image comparisons via joint-attention. Complemented by a progressive anti-shortcut training strategy, our approach guides models to capture subtle discriminative visual cues while preserving general instruction-following abilities. Extensive experiments demonstrate that \method achieves superior performance on both FGVC and VQA benchmarks, representing a practical approach for empowering MLLMs with fine-grained perception ability.

\noindent\textbf{Limitation and future work.} 
\label{sec:limitation}
While \method effectively leverages contrastive samples, our ablation study indicates that incorporating an excessive number of hard negatives can occasionally introduce context noise, which may slightly degrade performance on certain datasets like GLD. Future work could explore combining \method with recent data synthesis frameworks~\cite{jiao2025img}, which aims to deliver a fully automated training and self-improving pipeline for fine-grained multimodal perception.